\documentclass[sigconf]{acmart}

\renewcommand\footnotetextcopyrightpermission[1]{}
\usepackage{graphicx}
\usepackage{booktabs}
\usepackage{amsmath,amsfonts}
\usepackage{microtype}
\usepackage{xcolor}
\usepackage{enumitem}
\usepackage{tikz}
\usetikzlibrary{arrows.meta,positioning,fit,shapes.geometric,calc,shapes.misc}
\usepackage{caption}
\graphicspath{{figs/}}

\usepackage{graphicx}
\usepackage{booktabs}
\usepackage{caption}       
\usepackage{subcaption}    

\DeclareGraphicsExtensions{.pdf,.png,.jpg}
\graphicspath{{figs/}}

\setcitestyle{numbers,sort&compress}

\title{LLM-X: A Scalable Negotiation-Oriented Exchange for Communication Among Personal LLM Agents}

\author{Giuliano Lorenzoni}
\affiliation{%
  \institution{University of Waterloo}
  \city{Waterloo}
  \state{ON}
  \country{Canada}
}
\email{glorenzo@uwaterloo.ca}

\author{Paulo Alencar}
\affiliation{%
  \institution{University of Waterloo}
  \city{Waterloo}
  \state{ON}
  \country{Canada}
}
\email{palencar@uwaterloo.ca}

\author{Donald Cowan}
\affiliation{%
  \institution{University of Waterloo}
  \city{Waterloo}
  \state{ON}
  \country{Canada}
}
\email{dcowan@uwaterloo.ca}

\usepackage{listings}
\begin{document}






\begin{abstract}
We propose a \emph{personal-LLM exchange (LLM-X)}, a scalable negotiation oriented environment that enables direct, structured communication across populations of personal agents (LLMs), each representing an individual user. Unlike existing tool-centric protocols that focus on agent--API interaction, LLM-X introduces a message bus and routing substrate for \emph{LLM-to-LLM} coordination with guarantees around schema validity and policy enforcement.

We contribute: (1) an architecture for LLM-X comprising federated gateways, topic-based routing, and policy enforcement; (2) a typed message protocol supporting capability negotiation and contract-net--style coordination; and (3) the first empirical evaluation of LLM-based multi-agent negotiation at scale.

Experiments span 5, 9, and 12 agents, under distinct negotiation policies (Low, Medium, High), and across both short-run (minutes) and long-run (2h, 12h) load conditions. Results highlight clear policy--performance trade-offs: stricter policies improve robustness and fairness but increase latencies and message volume. Extended runs confirm that LLM-X remains stable under sustained load, with bounded latency drift.
\end{abstract}


\settopmatter{printacmref=false}
\renewcommand\footnotetextcopyrightpermission[1]{}
\pagestyle{plain}
\fancyhead{} 

\maketitle

\section{Introduction}

Large language models (LLMs) are increasingly deployed as personal assistants, but today in most cases they remain isolated: each user interacts with their own agent, with little or no direct cross-user communication. This isolation limits opportunities for negotiation, planning, and consensus, and also hinders systematic research on multi-agent dynamics under realistic scale. While prior work has studied multi-agent coordination in toy environments or through tool--API protocols, there is still no scalable substrate for structured \emph{agent--agent communication} among heterogeneous LLMs.

In this work we present the \textbf{LLM Exchange (LLM-X)}, a scalable negotiation-centric environment for structured cross-user communication. LLM-X introduces a message bus and routing substrate that enforces schema validity and policy rules, enabling direct typed negotiation while preserving fairness, safety, and consent. Unlike existing protocols centered on tool use, our design shifts the focus to \emph{agent--agent interaction}, providing a foundation for reproducible experiments at scale.

A key aspect of our design is that, although the current prototype instantiates agents as lightweight Python wrappers for controlled experimentation, the message model and negotiation protocols are LLM-ready. Each message envelope carries schema-validated JSON payloads that can originate from or be consumed by actual language models through APIs (e.g., OpenAI, Anthropic, HuggingFace). This ensures that the exchange is not merely a simulation of agent behavior, but a reusable substrate for coordinating heterogeneous personal LLM agents through typed negotiation. In practice, substituting scripted payloads with LLM-generated outputs preserves the same message flow, demonstrating that results obtained here directly reflect the dynamics of cross-LLM communication.

We evaluate LLM-X through controlled experiments with 5, 9, and 12 heterogeneous agents engaged in ContractNet-style negotiations. We systematically vary acceptance policies (low, medium, high) and measure trade-offs in latency, throughput, and communication overhead. Results highlight clear policy--performance trade-offs: stricter policies improve robustness and fairness but introduce higher latency and message volume. Extended 2h and 12h runs further confirm the environment’s stability, with bounded latency drift under sustained load.

\textbf{Contributions.} Taken together, our work provides:  
(i) an architecture for \emph{LLM-to-LLM exchange} with typed messages and policy enforcement;  
(ii) a reproducible evaluation of multi-agent negotiation based on communication efficiency, robustness, and fairness measures at scale; and  
(iii) empirical evidence of policy--performance trade-offs observed in realistic multi-agent populations.

\section{Background and Related Work}

Research on multi-agent systems with LLMs has expanded rapidly, highlighting challenges and opportunities across coordination, reasoning, and applications. Han et al.~\cite{han2025llm} provide a broad survey of open problems, emphasizing task allocation, reasoning through debates, layered context management, and memory design. Our work is complementary: while theirs outlines conceptual challenges, LLM-X contributes a concrete substrate and reproducible evaluation environment.

\textbf{Collaboration and Debate.} Recent studies have leveraged multi-agent debate or role-play to enhance reasoning and factuality~\cite{du2023debate,liang2024divergent,chan2023chateval,li2023camel}. Other frameworks such as \emph{AutoGen}~\cite{wu2023autogen}, \emph{MetaGPT}~\cite{hong2023metagpt}, and collaborative environments like \emph{Chatarena}~\cite{wu2023chatarena} showcase structured multi-agent interactions. These works highlight the potential of LLM-to-LLM communication but often lack a standardized communication substrate. LLM-X differs by introducing a schema-validated protocol with explicit policy controls, enabling systematic study of fairness, latency, and throughput.

\textbf{Planning and Task Decomposition.} Reasoning frameworks such as Chain-of-Thought (CoT)~\cite{wei2022cot}, Tree-of-Thoughts (ToT)~\cite{yao2023tot,long2023tot}, Graph-of-Thoughts (GoT)~\cite{besta2023got}, and Tab-CoT~\cite{ziqi2023tabcot} improve granularity of inference by decomposing problems into sub-steps. Other agent-oriented approaches—such as ReAct~\cite{yao2023react}, Reflexion~\cite{shinn2023reflexion}, Inner Monologue~\cite{huang2022inner}, and ReWOO~\cite{xu2023rewoo}—combine reasoning with environment feedback. While these methods operate mainly at the prompt or inference level, LLM-X abstracts the communication layer: our contribution is not a new reasoning algorithm but a substrate that can carry such reasoning outputs as structured negotiation payloads.

\textbf{Memory and Context.} Several works emphasize memory management for multi-agent systems, such as explicit memory-of-thoughts~\cite{li2024mot}, episodic and long-term memory design~\cite{wang2023memory}. Recent systems such as Reflexion~\cite{shinn2023reflexion}, Swiftsage~\cite{lin2023swiftsage}, and AgentBench~\cite{liu2023agentbench} explore self-reflection and context reuse. Retrieval-augmented agents~\cite{izacard2022atlas} further highlight the role of knowledge injection. These contributions address the internal state of agents. LLM-X complements them by focusing on the external communication protocol, ensuring memory traces remain schema-validated and comparable across experiments.

\textbf{Applications and Distributed Systems.} Practical applications include generative agents in interactive environments~\cite{park2023generative} and collective game-theoretic interaction~\cite{jinxin2023cgmi}. Recent agent frameworks—\emph{AutoGPT}~\cite{significant2023autogpt}, \emph{BabyAGI}~\cite{nakajima2023babyagi}, \emph{GPT-Engineer}~\cite{antonosika2023gptengineer}, and \emph{ScienceWorld}~\cite{wang2022scienceworld}—demonstrate the feasibility of distributed LLM-based systems in realistic tasks. These works showcase the promise of LLM-based multi-agent systems in realistic distributed settings. LLM-X aligns with this vision but stresses scalability and benchmarking: we provide controlled experiments (5–12 agents, varying policies) and extended runs (2 h, 12 h) to study negotiation under sustained load.

\textbf{Adaptation and Feedback.} Another complementary line of work focuses on improving LLM agents via instruction tuning and feedback alignment. Instruction-tuned models such as Flan-T5~\cite{chung2022flan}, Alpaca~\cite{taori2023alpaca}, and GPT-4-LLM~\cite{peng2023gpt4llm} democratized adaptation. RLHF and its extensions (\emph{InstructGPT}~\cite{ouyang2022instructgpt}, \emph{Constitutional AI}~\cite{bai2022constitutional}, \emph{DPO}~\cite{rafailov2023dpo}, and related datasets like \emph{Self-Instruct}~\cite{wang2023selfinstruct} and \emph{Unnatural Instructions}~\cite{honovich2022unnatural}) highlight how feedback loops refine agent behavior. While these efforts adapt models internally, LLM-X provides a complementary infrastructural contribution: a communication substrate that exposes adaptation and negotiation dynamics at the system level.

\textbf{Summary.} In sum, prior work has advanced debate, planning, memory, applications, and adaptation, but often lacks a shared substrate for communication. LLM-X fills this gap by contributing a reproducible, policy-aware exchange protocol that enables systematic experimentation with large populations of personal LLM agents. This situates our work as an infrastructural primitive for studying communication trade-offs at scale.

\section{System Overview: The LLM Exchange (LLM-X)}

Our design centers on an \emph{LLM Exchange (LLM-X)} that enables structured \emph{agent--agent communication} through negotiation protocols inspired by the classical \emph{Contract Net (CNet)} framework. Rather than relying on a generic message bus for pub/sub traffic, LLM-X focuses on typed negotiation rounds where one agent (the \emph{initiator}, e.g., Alice) issues a \emph{Call for Proposals (CFP)} and multiple \emph{contractors} respond with offers. Policy modules regulate acceptance and consensus formation, yielding tunable trade-offs between robustness, fairness, and efficiency. This protocol-compatible design borrows principles from standards such as MCP, but reorients the focus from tool--API interaction to direct cross-agent dialogue at scale.

\subsection{Architecture}

While our prototype agents are implemented in Python for controlled experimentation, the message envelope and negotiation protocols are designed to carry LLM-generated outputs (text or structured JSON) between heterogeneous models. This makes the exchange not just a Python object bus, but a reusable substrate for coordinating personal LLM agents. In practice, this ensures that the same infrastructure used in our experiments can integrate live LLMs via APIs (e.g., OpenAI, Anthropic, HuggingFace), with schema validation and policy enforcement applied uniformly regardless of the underlying model.

\begin{figure*}[t]
  \centering
  \resizebox{\textwidth}{!}{\includegraphics{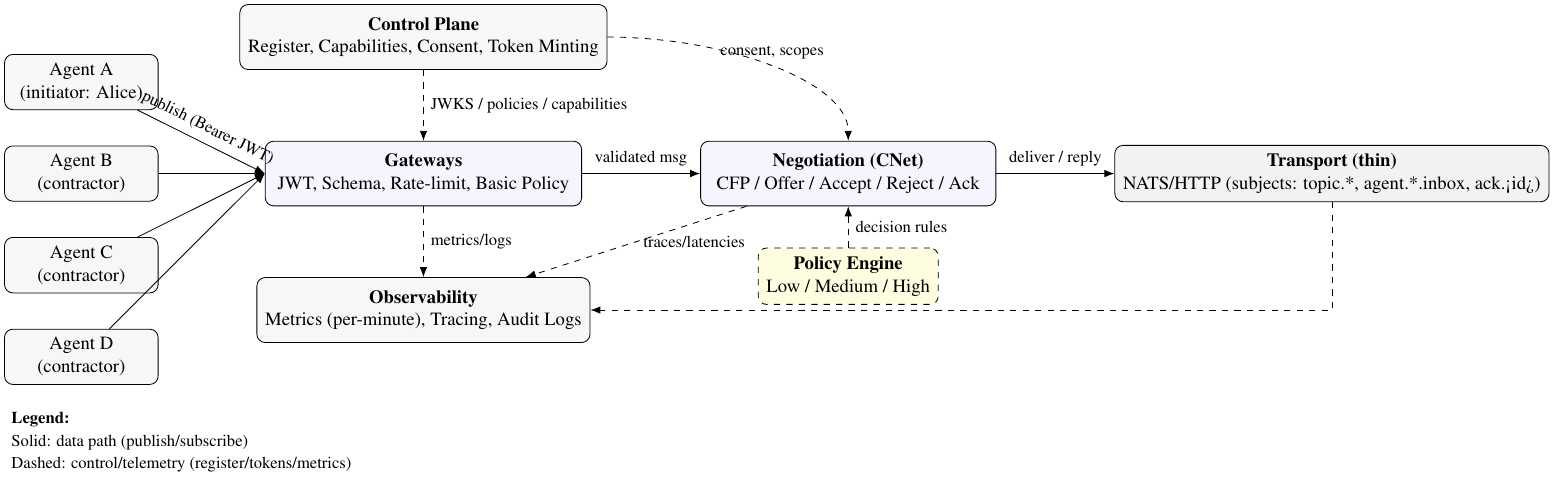}}
  \caption{LLM-X conceptual (negotiation-centric) architecture.}
  \label{fig:llmx-architecture}
\end{figure*}

Figure~\ref{fig:llmx-architecture} illustrates the high-level architecture. Agents connect via lightweight gateways that enforce authentication (JWT), schema validation, and basic rate limiting. Negotiation proceeds through structured message types (\texttt{CFP}, \texttt{Offer}, \texttt{Accept}, \texttt{Reject}, \texttt{Ack}). A \textbf{Policy Engine} enforces rules (Low, Medium, High) that govern how Alice, as initiator, selects among competing offers. The \textbf{Control Plane} handles agent registration, consent, and capability negotiation, while \textbf{Observability} components capture detailed traces, metrics, and audits for later analysis.

\subsection{Negotiation Flow}

\begin{figure*}[t]
  \centering
  \resizebox{\textwidth}{!}{\includegraphics{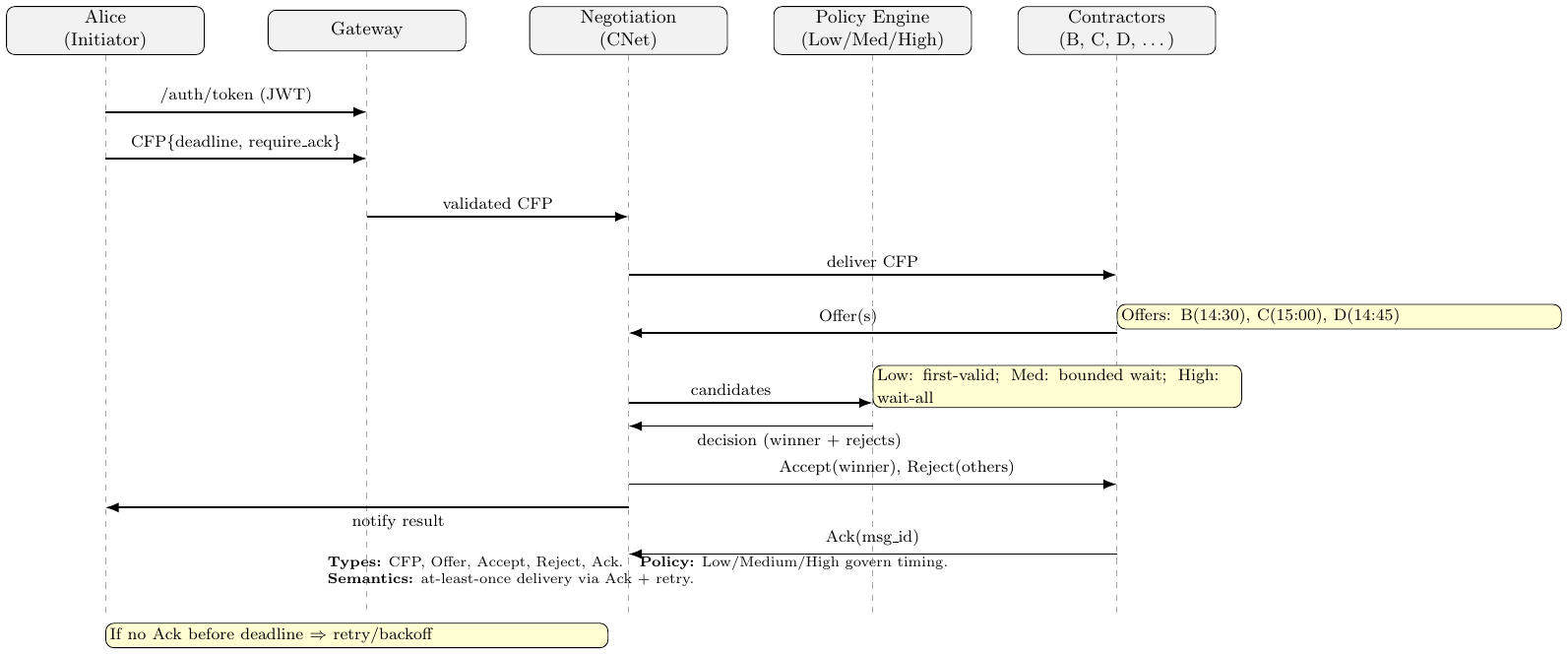}}
  \caption{CNet-style conceptual negotiation sequence in LLM-X. Alice emits a \texttt{CFP} with a deadline; contractors reply with \texttt{Offer}s. The \textbf{Policy Engine} governs when and how the winner is selected (Low: first-valid; Medium: bounded wait; High: wait-all). Acks ensure reliable delivery; retries occur on missing acknowledgments.}
  \label{fig:llmx-sequence}
\end{figure*}

Figure~\ref{fig:llmx-sequence} shows a typical negotiation sequence. Alice (initiator) publishes a \texttt{CFP} with a deadline; contractors respond with \texttt{Offers}. The Policy Engine applies rules: under \textbf{Low}, the first valid offer is accepted; under \textbf{Medium}, Alice waits for a bounded set of replies before selecting; under \textbf{High}, all offers must be collected before decision. Contractors receive either \texttt{Accept} or \texttt{Reject} messages, and \texttt{Ack} ensures delivery reliability. Retries are triggered if acknowledgments fail.

\noindent
Each ContractNet-style message in LLM-X is encapsulated in a typed JSON \emph{envelope} carrying metadata such as \texttt{msg\_id}, \texttt{from/to}, \texttt{capabilities}, and \texttt{scope} (with TTLs and consent). Payloads reference explicit JSON Schemas (e.g., \texttt{Negotiation.json}) that define the structure of \texttt{CFP}, \texttt{Offer}, and \texttt{Accept/Reject} acts, enabling validation at the gateway. Transport is realized over NATS \emph{subjects} (e.g., \texttt{topic.*}, \texttt{agent.*.inbox}, \texttt{ack.<id>}), providing pub/sub delivery and reliability via acknowledgments and retries. This combination of envelope, schema, and transport substrate bridges the abstract negotiation flow with an enforceable protocol, and naturally motivates the subsequent discussion of FIPA-style alternating offers (Section~3.3) and the detailed message model (Section~3.4).

\subsection{FIPA Alternating Offers}

Beyond the ContractNet-style protocol, LLM-X also supports negotiation flows compatible with the \emph{FIPA Alternating Offers} standard. In this setting, two agents iteratively exchange counter-proposals under explicit deadlines until either agreement or termination. We implement this pattern by encoding each round in a JSON \emph{Offer} payload referencing a dedicated schema (\texttt{AlternatingOffer.json}), with fields for proposed terms, validity intervals, and acceptance conditions. Gateways validate each counter-proposal against the schema and enforce policy constraints (e.g., deadlines or fairness rules), while NATS subjects handle message routing and retries. This extension demonstrates that LLM-X is not tied to a single coordination primitive but can accommodate a spectrum of agent communication protocols. Complete schema sketches and implementation notes are provided in Appendix~\ref{appendix:schemas}.

\subsection{Message Model and Protocol}

At the core of LLM-X is a minimal, typed JSON envelope that binds each message to an explicit schema (for CFP, Offer, Accept/Reject, Ack) and to a consent/scope context. This guarantees \emph{interoperability} (agents agree on types) and \emph{safety} (schemas are validated at the gateway before delivery), while remaining \emph{LLM-grounded}: the same envelope can be produced/consumed by live LLM APIs without changing the negotiation flow. Three delivery modes are supported—direct (inbox), topic (broadcast within a negotiation), and group (targeted subsets with QoS)—and policy modules (Low/Medium/High) regulate decision timing and completeness.

For space, we move concrete schema sketches and the full envelope example to Appendix~\ref{appendix:schemas}, and transport/JWT/rate-limit details to Appendix~\ref{appendix:transport}. This preserves the key claim (typed, policy-aware, LLM-ready exchange) while deferring engineering specifics.

\subsection{Security and Privacy}

Identity is linked to each user agent, with consent required per negotiation. Sensitive fields are sanitized, and logs retain audit trails. Policy enforcement prevents acceptance of malformed or duplicate offers. Threats such as prompt injection are mitigated by schema validation and redaction of sensitive payloads.

\section{Environment and Scaling Experiments}
\label{sec:environment}

\subsection{Prototype Setup}
All experiments were conducted using Python-based agents configured with a common communication harness. Each agent connects through a lightweight gateway enforcing JWT-based authentication, JSON Schema validation, and rate limiting. The negotiation layer follows a ContractNet-style protocol (CFP $\rightarrow$ Offer $\rightarrow$ Accept/Reject), regulated by a Policy Engine. We evaluate three distinct acceptance policies: a \textbf{Low} policy, which closes early after a timeout at the risk of missing a final response; a \textbf{Medium} policy, which terminates as soon as the first valid response arrives; and a \textbf{High} policy, which waits to collect all offers before making a decision.

Agent populations include 5, 9, and 12 contractors, with Alice acting as the initiator. Load is generated via \texttt{traffic\_gen} using a Poisson arrival process to simulate realistic, stochastic interaction patterns.

\subsection{Metrics}
We capture both system-level and communication-level metrics, including \textbf{per-minute volumes} (counts of CFPs and Offers), \textbf{latencies} measured as median (p50) and tail (p95) values across rounds, and \textbf{per-round offers} capturing the distribution of received Offers. All metrics are logged continuously and aggregated into CSV and JSONL traces for post-analysis.

\subsection{Experimental Design}
\label{sec:design}
Our evaluation is structured around controlled experiments that vary the number of agents, acceptance policies, and run duration. All agents are implemented in Python and communicate via the LLM-X substrate, with Alice acting as initiator and a configurable set of contractors responding.

\paragraph{Policies.}
The \textbf{Low} policy closes on the first counter or times out, risking no final response. The \textbf{Medium} policy accepts the first valid response and closes immediately. The \textbf{High} policy requires Alice to wait until multiple offers are collected before applying the decision rules defined by the Policy Engine.

\paragraph{Agent sets.}
We study populations of \{5, 9, 12\} contractors in addition to Alice as initiator. In our prototype, these contractors are instantiated as Python wrappers for reproducibility, but the protocol itself is agnostic: payloads could equally be produced by live LLMs. Since all message types are schema-validated JSON, integration with APIs such as OpenAI, Anthropic, or Hugging Face requires no modification to the negotiation flow.

\paragraph{Load generation.}
Traffic is generated using a Poisson process (\texttt{traffic\_gen}) to emit \texttt{CFP} (Call-for-Proposal) messages at a specified arrival rate. Contractors produce Offers, and responses are governed by the active policy.

\paragraph{Metrics.}
For each scenario we capture per-minute counts of CFPs and Offers, per-round offer distributions, and latencies at both the p50 and p95 levels across all messages. Additionally, aggregate throughput and sustained stability are monitored to detect potential drift over long runs.

\paragraph{Scenarios.}
Our experiments comprise three main scenarios: 
(1) \textbf{Scaling short runs}, evaluating 5, 9, and 12 agents under Low, Medium, and High policies (2-minute runs); 
(2) \textbf{Extended runs (Policy High)}, performing 2-hour tests with 5, 9, and 12 agents; and 
(3) \textbf{Extended runs (12 agents)}, including 2-hour Medium, 2-hour High, and 12-hour High configurations.

Each short-run scenario (2 minutes) was repeated three times under independent random seeds. The objective of these repetitions was not formal statistical significance but to confirm consistency and reproducibility of observed trends. Across runs, variability in per-minute volumes and latency metrics remained negligible, confirming that the system produces stable patterns under identical configurations.

\paragraph{Warm-up runs.}
Prior to the main experiments, short preliminary runs were executed with 5 agents under Policy Low (2 minutes, cumulative and final). These validated schema correctness, logging, and environment stability. As their purpose was setup verification rather than performance evaluation, they are not analyzed in detail in Section~\ref{sec:results}.







\section{Results}
\label{sec:results}

We now present results from both the warm-up runs and the main evaluation scenarios.
Results are organized into three parts: (i) warm-up validation, (ii) scaling with the
number of agents and policy strictness, and (iii) extended load experiments.

\subsection{Warm-up Validation (5 Agents, Policy Low)}

We first ran short 2-minute tests to validate schema, logging, and negotiation stability. 
\textbf{Test~1} (3 CFPs, 15 Offers) showed communication but no complete rounds. 
\textbf{Test~2} (6 CFPs, 30 Offers) captured latency between 2--8\,ms with full rounds. 
\textbf{Test~3} (3 CFPs, 15 Offers) produced stable latencies of 1--3\,ms and consistent logs. 
This progression confirmed the setup’s readiness for long-duration experiments.

\subsection{Scaling with Number of Agents}
\label{sec:scaling}

\begin{figure}[!t]
  \centering
  \begin{minipage}[t]{0.31\linewidth}
    \centering
    \includegraphics[width=\linewidth]{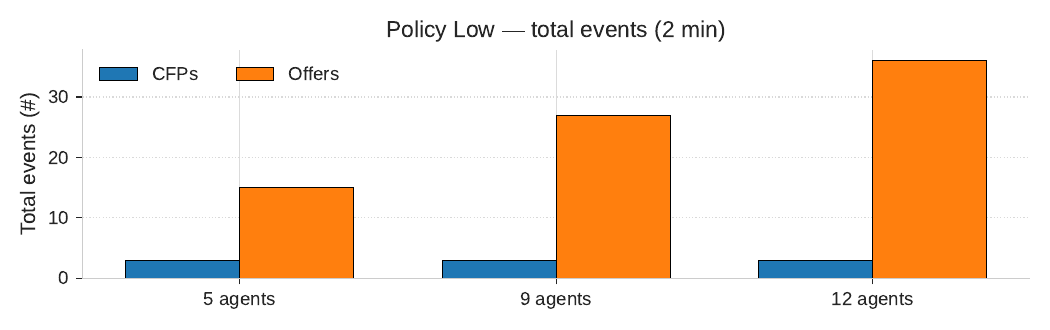}
    \vspace{2pt}
    {\scriptsize (a) Total events\par}
  \end{minipage}\hfill
  \begin{minipage}[t]{0.31\linewidth}
    \centering
    \includegraphics[width=\linewidth]{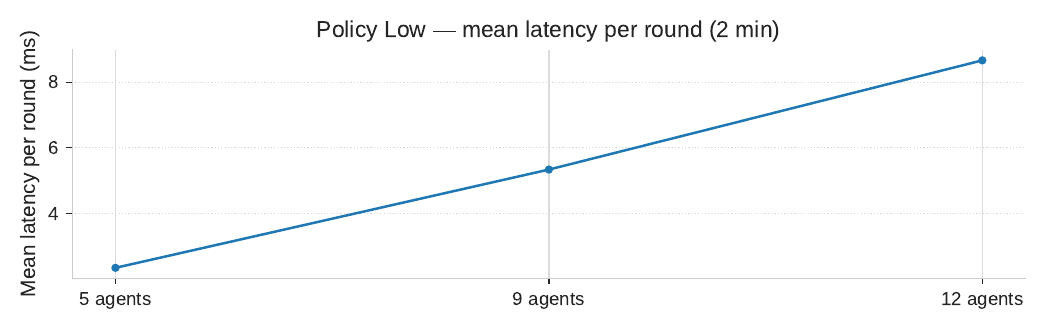}
    \vspace{2pt}
    {\scriptsize (b) Mean latency per round\par}
  \end{minipage}\hfill
  \begin{minipage}[t]{0.31\linewidth}
    \centering
    \includegraphics[width=\linewidth]{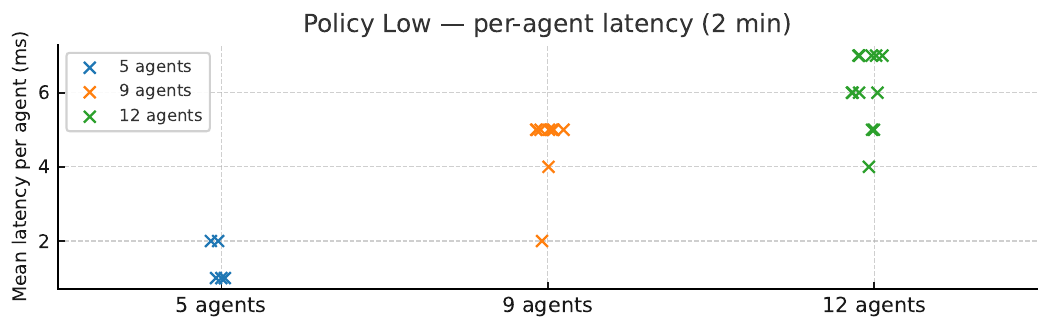}
    \vspace{2pt}
    {\scriptsize (c) Mean latency per agent\par}
  \end{minipage}
  \vspace{2pt}
  \caption{Consolidated results for \textbf{Low Acceptance Policy} (2 min), comparing 5, 9, and 12 agents.}
  \label{fig:policy_low_compact}
\end{figure}

We analyze 2-minute runs under the \textbf{Low} policy with 5, 9, and 12 agents, measuring CFP/Offer volumes, round completeness, and latencies (Table~\ref{tab:cmp_low2min}, Figure~\ref{fig:policy_low_compact}).  

With \textbf{5 agents}, 3 CFPs produced 15 offers (5 per round), all rounds completed, and latencies remained extremely low (1--3\,ms; p95 $\sim$2--3\,ms).  
At \textbf{9 agents}, the same pattern held with 27 offers (9 per round), though mean latency rose to 4--5\,ms (p95 7--9\,ms).  
At \textbf{12 agents}, 36 offers (12 per round) were observed, but latency increased further (5--7\,ms mean; p95 8--11\,ms) with greater dispersion across agents.  

Overall, the \textbf{Low policy} scales linearly in offer counts and preserves round completeness across agent numbers, but latency and heterogeneity gradually increase with concurrency. This illustrates the \textbf{trade-off between scalability and temporal consistency}: the protocol remains robust, yet synchronization becomes less uniform as agent populations grow.

\begin{table}[!t]
\centering
\caption{Consolidated comparison — Low Policy (2 min), 5 vs 9 vs 12 agents}
\label{tab:cmp_low2min}
\resizebox{\linewidth}{!}{%
\begin{tabular}{llll}
\toprule
                        Aspect &                                        5 agents &                                        9 agents &                                       12 agents \\
\midrule
           Identified agents   &                                              5 &                                              9 &                                             12 \\
                       Policy  &                                            Low &                                            Low &                                            Low \\
 Duration (minutes recorded)   &                                 2 (00:06--00:07) &                                 3 (04:31--04:33) &                                 2 (05:25--05:26) \\
                 CFPs issued   &                                              3 &                                              3 &                                              3 \\
              Offers received  &                                             15 &                                             27 &                                             36 \\
            Rounds recorded    &                                              3 &                                              3 &                                              3 \\
           Offers per round    &                                       always 5 &                                       always 9 &                                      always 12 \\
 Approx. mean latency (ms)     &                                          1--3 &                                          4--5 &                                          5--7 \\
 Approx. p95 latency (ms)      &                                          2--3 &                                          7--9 &                                         8--11 \\
   Conversation termination    & Completed after last offer (no explicit ``accept'') & Completed after last offer (no explicit ``accept'') & Completed after last offer (no explicit ``accept'') \\
\bottomrule
\end{tabular}%
}
\end{table}

\subsection{Policy Comparisons}
\label{sec:policy-comparisons}
Direct comparison between Medium and High acceptance policies, highlighting trade-offs:
speed versus completeness, and differences in message overhead.

\subsubsection{Medium Acceptance Policy}

\begin{figure}[!t]
  \centering
  \begin{minipage}[t]{0.31\linewidth}
    \centering
    \includegraphics[width=\linewidth]{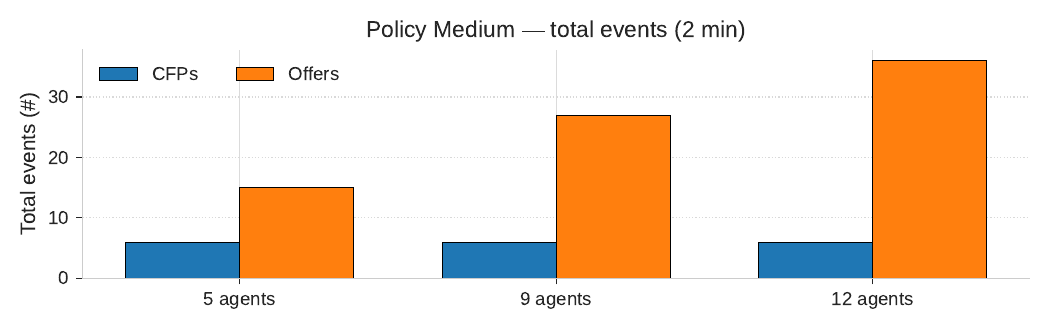}
    \vspace{2pt}
    {\scriptsize (a) Total events\par}
  \end{minipage}\hfill
  \begin{minipage}[t]{0.31\linewidth}
    \centering
    \includegraphics[width=\linewidth]{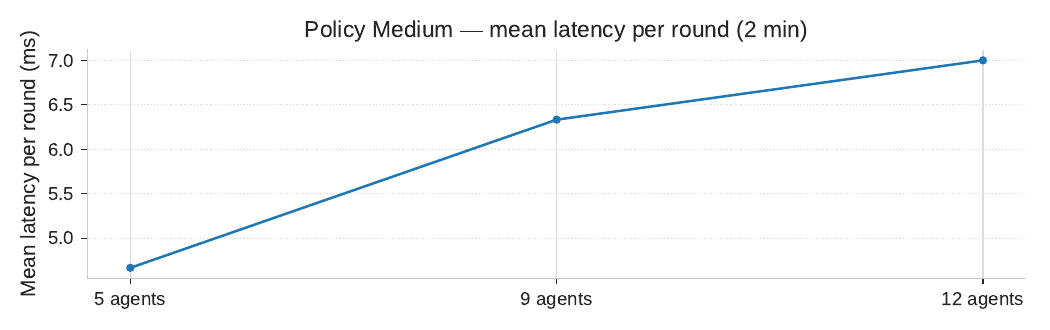}
    \vspace{2pt}
    {\scriptsize (b) Mean latency per round\par}
  \end{minipage}\hfill
  \begin{minipage}[t]{0.31\linewidth}
    \centering
    \includegraphics[width=\linewidth]{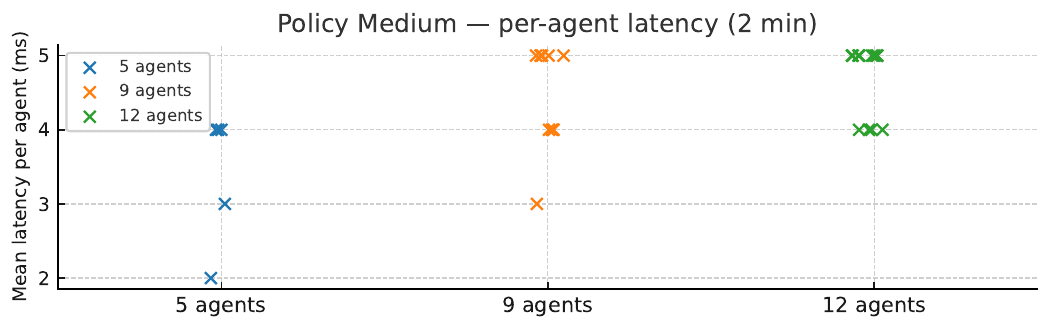}
    \vspace{2pt}
    {\scriptsize (c) Mean latency per agent\par}
  \end{minipage}
  \vspace{2pt}
  \caption{Consolidated results for \textbf{Medium Acceptance Policy} (2 min), comparing 5, 9, and 12 agents.}
  \label{fig:policy_med_compact}
\end{figure}

We analyze short 2-minute runs under the \textbf{Medium} policy with 5, 9, and 12 agents. Unlike Low, Medium introduces an explicit \texttt{Confirm}, allowing rounds to terminate once the first valid offer is accepted. This accelerates resolution but may end negotiations before all offers are collected. Table~\ref{tab:cmp_med2min} and Figure~\ref{fig:policy_med_compact} summarize the comparison.

With \textbf{5 agents}, 6 CFPs yielded 30 Offers and 3 Confirms, with mean latencies of \textbf{3--4 ms} and p95 of \textbf{6--7 ms}. For \textbf{9 agents}, 27 Offers were observed with similar early terminations; latencies increased to \textbf{5 ms} (p95: \textbf{8--9 ms}). At \textbf{12 agents}, 36 Offers were recorded, again with some premature closures, and latencies averaged \textbf{4--6 ms} (p95: \textbf{7--8 ms}). 

Overall, the \textbf{Medium} policy maintains scalability but shifts negotiation dynamics: faster decisions occur at the cost of completeness, highlighting the trade-off between speed and full participation.

\begin{table}[!t]
\centering
\caption{Consolidated comparison — Medium Acceptance Policy (2 min), 5 vs 9 vs 12 agents}
\label{tab:cmp_med2min}
\resizebox{\linewidth}{!}{%
\begin{tabular}{llll}
\toprule
                        Aspect &                                        5 agents &                                        9 agents &                                       12 agents \\
\midrule
           Identified agents   &                                              5 &                                              9 &                                             12 \\
                       Policy  &                                         Medium &                                         Medium &                                         Medium \\
 Duration (minutes recorded)   &                                              1 &                                              1 &                                              1 \\
                 CFPs issued   &                                              6 &                                              6 &                                              6 \\
              Offers received  &                                             30 &                                             27 &                                             36 \\
             Confirms received &                                              3 &                                              3 &                                              3 \\
            Rounds recorded    &                                              6 &                                              6 &                                              6 \\
           Offers per round    &                        5 when complete; some terminated early via Confirm & 9 when complete; some terminated early via Confirm & 12 when complete; some terminated early via Confirm \\
 Approx. mean latency (ms)     &                                          3--4 &                                              5 &                                          4--6 \\
 Approx. p95 latency (ms)      &                                          6--7 &                                          8--9 &                                          7--8 \\
   Conversation termination    & Some rounds ended via Confirm, others complete & Some rounds ended via Confirm, others complete & Some rounds ended via Confirm, others complete \\
\bottomrule
\end{tabular}%
}
\end{table}

\subsubsection{High Acceptance Policy}

\begin{figure}[!t]
  \centering
  \begin{minipage}[t]{0.31\linewidth}
    \centering
    \includegraphics[width=\linewidth]{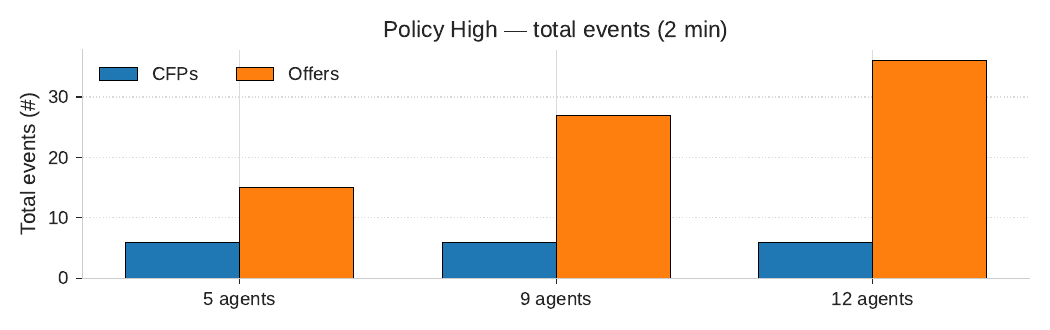}
    \vspace{2pt}
    {\scriptsize (a) Total events\par}
  \end{minipage}\hfill
  \begin{minipage}[t]{0.31\linewidth}
    \centering
    \includegraphics[width=\linewidth]{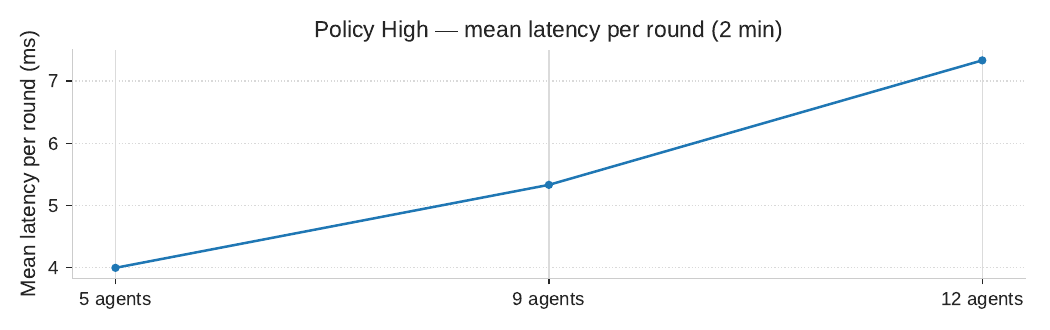}
    \vspace{2pt}
    {\scriptsize (b) Mean latency per round\par}
  \end{minipage}\hfill
  \begin{minipage}[t]{0.31\linewidth}
    \centering
    \includegraphics[width=\linewidth]{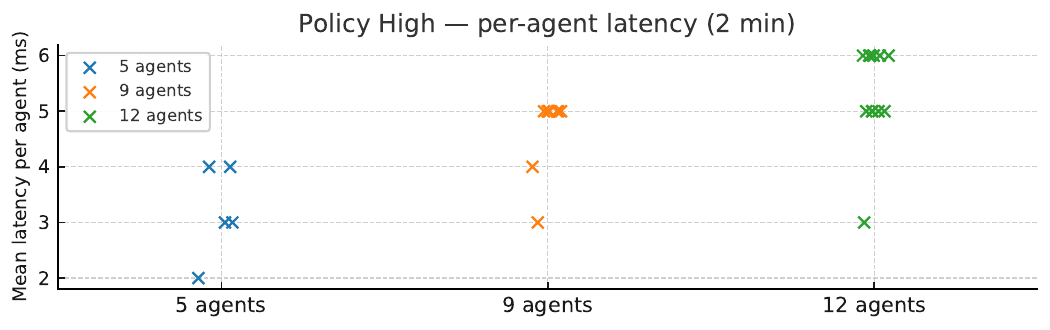}
    \vspace{2pt}
    {\scriptsize (c) Mean latency per agent\par}
  \end{minipage}
  \vspace{2pt}
  \caption{Consolidated results for \textbf{High Acceptance Policy} (2 min), comparing 5, 9, and 12 agents.}
  \label{fig:policy_high_compact}
\end{figure}

Table~\ref{tab:policy_high} and Figure~\ref{fig:policy_high_compact} summarize short 2-minute runs with 5, 9, and 12 agents under the \textbf{High} policy. Unlike Medium, all offers are collected before decisions, ensuring complete rounds but increasing overhead.

With \textbf{5 agents}, 3 CFPs produced 15 Offers (5 per round). Latencies remained low, with per-agent means of \textbf{3 ms} and p95 of \textbf{5–6 ms}.  
At \textbf{9 agents}, the same structure held (27 Offers from 3 CFPs), with slightly higher latencies (\textbf{4–5 ms} mean, \textbf{7–8 ms} p95).  
For \textbf{12 agents}, 6 CFPs alternated between full and empty rounds (12–0–12–0–12–0), yielding 3 effective rounds. Mean per-agent latency rose to \textbf{5–6 ms} (p95 \textbf{8–9 ms}), with per-round averages near \textbf{7 ms}.  

Overall, the High policy guarantees offer completeness and bounded latencies in the millisecond range, but synchronization becomes less uniform at higher concurrency, as seen in the alternating pattern with 12 agents.

\begin{table}[!t]
\centering
\caption{High Acceptance Policy – 2-minute runs with 5, 9, and 12 agents. Metrics include CFP and Offer volumes, latency distributions, and round completeness.}
\label{tab:policy_high}
\resizebox{\linewidth}{!}{%
\begin{tabular}{lccc}
\toprule
\textbf{Metric} & \textbf{5 Agents} & \textbf{9 Agents} & \textbf{12 Agents} \\
\midrule
CFPs issued & 3 & 3 & 6 \\
Offers received & 15 & 27 & 36 \\
Confirms & 0 & 0 & 0 \\
Accepts & 0 & 0 & 0 \\
Rejects & 0 & 0 & 0 \\
Rounds registered & 3 & 3 & 6 (3 effective + 3 empty) \\
Offers per round & always 5 & always 9 & 12/0 alternating \\
Mean latency per agent (ms) & 3.2 & 4.7 & 5.3 \\
p95 latency per agent (ms) & 5.4 & 7.3 & 8.3 \\
Mean latency per round (ms) & 4.1 & 6.2 & 7.3 \\
Conversation termination & Completed (no Accepts) & Completed (no Accepts) & Completed (no Accepts) \\
\bottomrule
\end{tabular}%
}
\end{table}

\subsection{Extended Load Experiments}
We evaluate longer scenarios: 2h runs under \textbf{Medium} and \textbf{High} policies to assess stability under sustained load, and a 12h run with \textbf{High} (12 agents) as a stress test for latency drift and throughput stability.

\subsubsection{Extended 2-Hour Runs (12 Agents)}

We now assess stability under sustained load and long-horizon effects (e.g., latency drift, throughput saturation, robustness). Following the short-run analysis in Sections~\ref{sec:scaling}--\ref{sec:policy-comparisons}, we focus on \textbf{12 agents} and only the \textbf{Medium} and \textbf{High} policies. This choice reflects three factors: (i) 12 agents are the most demanding concurrency tested, (ii) Medium and High explicitly regulate decision timing and completeness, and (iii) Low, with its early-close behavior and no explicit confirmation, offers limited diagnostic value over long horizons (as also seen in the short-run results).\footnote{See Sections~\ref{sec:scaling} and~\ref{sec:policy-comparisons} for the 2-minute analyses across 5/9/12 agents under Low/Medium/High.}  

\begin{figure}[!t]
  \centering
  \begin{minipage}[t]{0.31\linewidth}
    \centering
    \includegraphics[width=\linewidth]{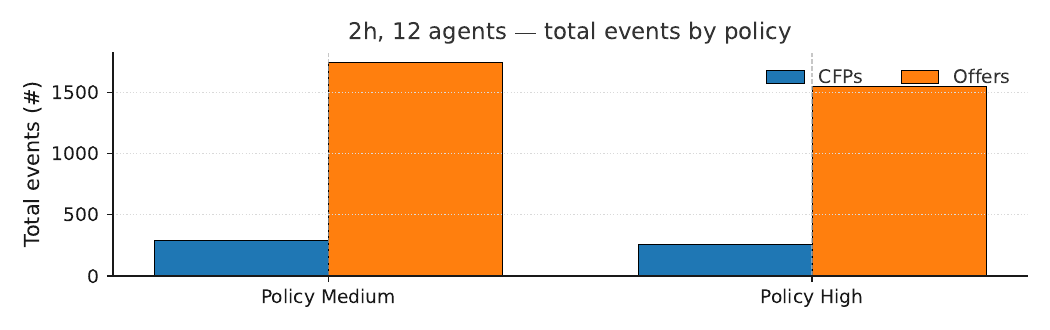}
    \vspace{2pt}
    {\scriptsize (a) Total events\par}
  \end{minipage}\hfill
  \begin{minipage}[t]{0.31\linewidth}
    \centering
    \includegraphics[width=\linewidth]{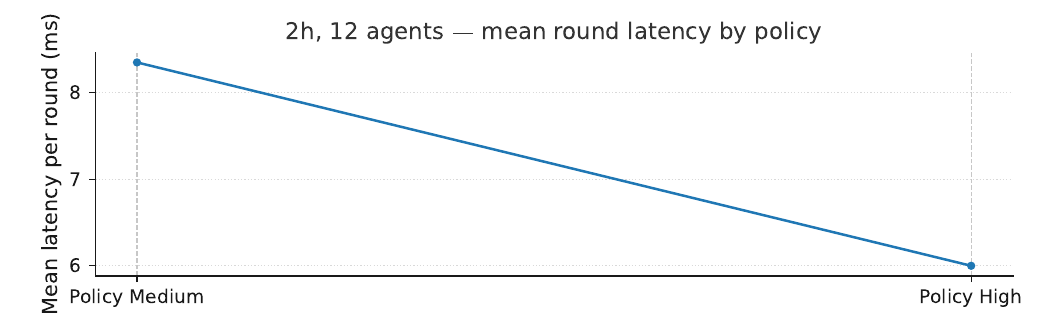}
    \vspace{2pt}
    {\scriptsize (b) Mean latency per round\par}
  \end{minipage}\hfill
  \begin{minipage}[t]{0.31\linewidth}
    \centering
    \includegraphics[width=\linewidth]{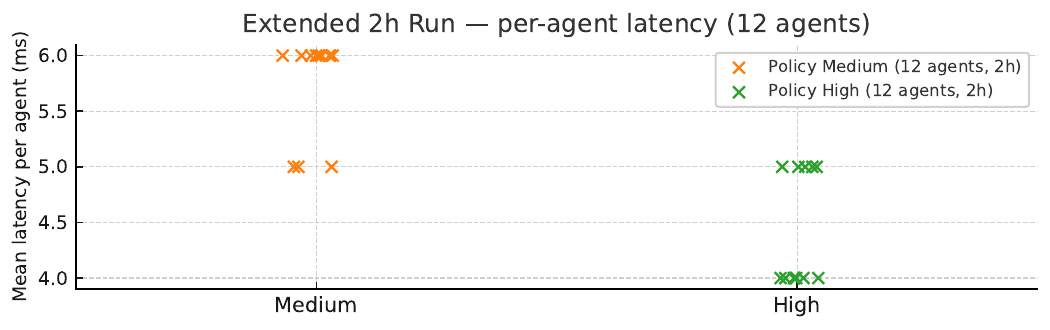}
    \vspace{2pt}
    {\scriptsize (c) Mean latency per agent\par}
  \end{minipage}
  \vspace{2pt}
  \caption{Extended runs (2h, 12 agents): \textbf{Policy Medium} vs \textbf{Policy High}.}
  \label{fig:ext_12agents_2h_compact}
\end{figure}

In the 2-hour runs a clear trade-off emerges (Table~\ref{tab:ext_2h_med_high}, Figure~\ref{fig:ext_12agents_2h_compact}). \textbf{Medium} sustains higher throughput—more CFPs, offers, and completed rounds—while still achieving 100\% completion, but with slightly higher latencies (mean $\sim$8.3\,ms, p95 12\,ms). \textbf{High} enforces stricter synchronization, yielding fewer rounds and messages but consistently lower and more stable latencies (mean $\sim$6.0\,ms, p95 10\,ms). Thus, Medium maximizes message volume and activity, whereas High prioritizes latency reduction and temporal stability under sustained load.

\begin{table}[!t]
  \centering
  \includegraphics[width=0.66\linewidth]{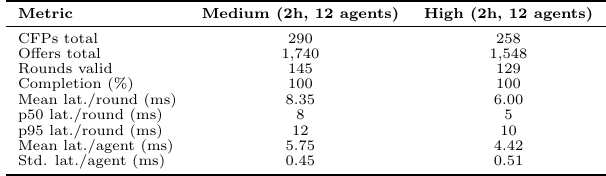}
  \caption{Extended runs (2h, 12 agents): comparison between \textbf{Medium} and \textbf{High} policies.}
  \label{tab:ext_2h_med_high}
\end{table}

\subsubsection{Extended 12-Hour Run (12 Agents)}
\label{sec:res_ext12h}

\begin{figure*}[t]
  \centering
  \begin{minipage}[t]{0.31\linewidth}
    \centering
    \includegraphics[width=\linewidth]{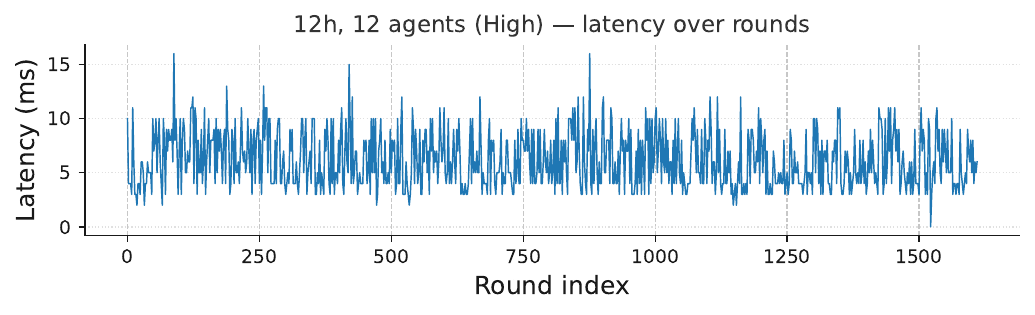}
    \vspace{2pt}
    {\scriptsize (a) Latency over time (12 h)\par}
  \end{minipage}\hfill
  \begin{minipage}[t]{0.31\linewidth}
    \centering
    \includegraphics[width=\linewidth]{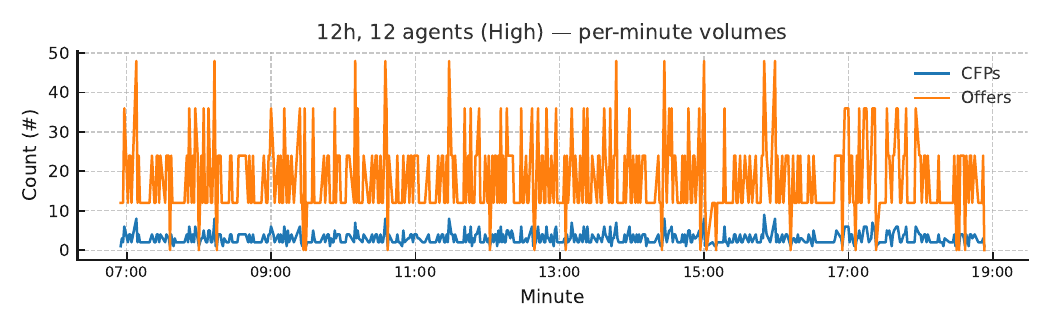}
    \vspace{2pt}
    {\scriptsize (b) Per-minute CFPs and Offers\par}
  \end{minipage}\hfill
  \begin{minipage}[t]{0.31\linewidth}
    \centering
    \includegraphics[width=\linewidth]{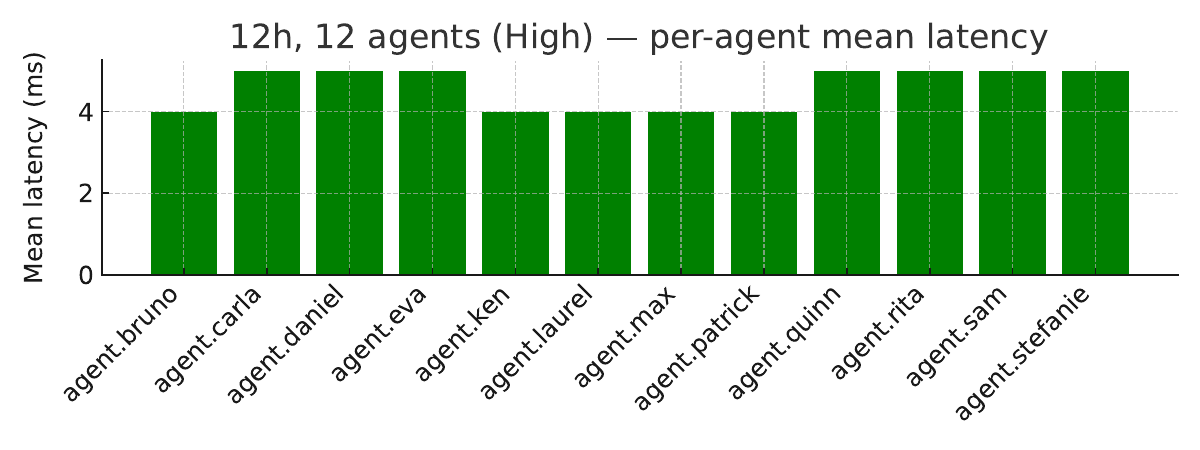}
    \vspace{2pt}
    {\scriptsize (c) Per-agent mean latency\par}
  \end{minipage}

  \vspace{4pt}
  \caption{Extended 12-hour experiment with \textbf{12 agents, High Policy}.
  Results show latency stability over time, sustained per-minute traffic,
  and homogeneous latency distribution across agents.}
  \label{fig:ext_high12_12h_trio}
\end{figure*}

The 12-hour experiment with 12 agents under the \textbf{High Policy} sustained
\emph{100\% round completeness} (806 valid rounds) and produced 1{,}612 CFPs and
9{,}672 offers (Table~\ref{tab:ext_12h_high12}). Latencies remained consistently
low and stable, with a mean of 6.18\,ms (p50 = 5\,ms, p95 = 11\,ms), and
homogeneous per-agent averages (mean 4.58\,ms, std.\ dev.\ $\approx$ 0.51\,ms),
confirming uniform behavior across agents. Per-minute traces showed steady,
regular traffic without saturation peaks, and latency trends exhibited no
significant drift across the full 12-hour horizon (Figure~\ref{fig:ext_high12_12h_trio}).
These results confirm that \textbf{High Policy} maintains \emph{stable low-latency
performance} and temporal stability over prolonged durations, albeit at a slower
round and message pace than the \textbf{Medium Policy}, consistent with the trade-offs
observed in the 2-hour runs. A detailed quantitative breakdown of the 12-hour run is
provided in Appendix~\ref{appendix:transport} (Table~\ref{tab:ext_12h_high12},
Figure~\ref{fig:ext_high12_12h_trio}).

\begin{table}[t]
  \centering
  \caption{Extended 12-hour run with 12 agents under the \textbf{High}
  acceptance policy. Results show total rounds, messages, and latency statistics.}
  \label{tab:ext_12h_high12}
  \resizebox{\linewidth}{!}{%
    \begin{tabular}{lcccccc}
      \toprule
      \textbf{Policy} & \textbf{Rounds} & \textbf{CFPs} & \textbf{Offers} &
      \textbf{Mean Lat. (ms)} & \textbf{p50 Lat. (ms)} & \textbf{p95 Lat. (ms)} \\
      \midrule
      High (12 agents, 12 h) & 806 & 1{,}612 & 9{,}672 & 6.18 & 5 & 11 \\
      \bottomrule
    \end{tabular}%
  }
\end{table}

\section{Discussion}
\label{sec:discussion}

\subsection{Engineering and Architectural Considerations}
From an engineering perspective, LLM-X was intentionally designed as a 
plug-and-play substrate for multi-agent negotiation. Although the present 
experiments use synthetic Python agents, the architecture is already compatible 
with live LLM endpoints (e.g., OpenAI, Anthropic, HuggingFace) via the gateway’s 
schema-validated JSON interface. No protocol or API changes are required to 
replace synthetic payloads with real model outputs, ensuring future-proof 
interoperability. Crucially, stochastic LLM outputs are handled through schema validation, automated retry and correction mechanisms, and policy enforcement at the gateway level, enabling safe integration without altering the experimental setup.

\subsection{System Design and Shielding Measures}
The gateway layer acts as the first engineering shield, enforcing 
JWT-based authentication, JSON Schema validation, and concurrency-safe queuing. 
These mechanisms guarantee that message exchange remains stable and auditable 
even under stochastic load. Observability hooks enable deterministic replay and 
trace-level inspection, establishing a foundation for explainable and governed 
multi-agent behavior.

A second layer of protection lies in the Policy Abstraction Layer, which 
separates decision logic from communication routines. This abstraction allows 
dynamic switching between acceptance policies (Low, Medium, High) without 
recompilation, ensuring that negotiation fairness and responsiveness can be 
tuned per deployment scenario.

\subsection{Scalability Lessons}
The experiments show that increasing agent populations from 5 to 12 scales 
message traffic nearly linearly, while latency degradation remains bounded. 
Throughput stability across 2-hour and 12-hour runs suggests that the asynchronous 
gateway is CPU-bound rather than I/O-bound, an important insight for distributed 
deployments. Future iterations may employ multiprocessing or gRPC streaming to 
improve concurrency and further stress-test performance.

\subsection{Future Engineering Extensions}
Next steps include containerized deployment for reproducible benchmarks, 
integration with live LLM APIs to validate real negotiation flows, and 
reinforcement-driven tuning of policy parameters. Together, these extensions will 
transition LLM-X from a prototype substrate into a fully governed agentic testbed 
for negotiation, coordination, and long-horizon decision assurance.

\section{Conclusions}
\label{sec:conclusion}

This paper introduced LLM-X, a scalable negotiation-oriented exchange 
for communication among personal LLM agents. Through controlled multi-agent 
experiments under varying acceptance policies and time horizons, we demonstrated 
that LLM-X maintains stable throughput and latency while preserving fairness and 
round completeness. The 12-hour runs confirmed that agent synchronization and 
protocol efficiency persist over long horizons, highlighting the architecture’s 
resilience to drift and saturation.

From an engineering standpoint, the system’s plug-and-play design, schema-validated 
gateway, and policy abstraction layer form a reproducible substrate for governed 
multi-agent negotiation. These components ensure that the framework is immediately 
extendable to live LLM endpoints, providing a \textbf{safe, observable, and 
scalable foundation} for studying coordination and reasoning among autonomous 
language models.

Future work will integrate real LLMs and reinforcement-driven policy control, 
transitioning LLM-X from simulation to a deployable testbed for agentic 
decision assurance. This trajectory connects experimental reproducibility with 
real-world readiness—bridging engineering reliability and cognitive adaptability 
in next-generation multi-agent systems.


\bibliographystyle{ACM-Reference-Format}
\bibliography{references} 

\appendix

\section{Message Schemas (Sketches)}
\label{appendix:schemas}

At the core of LLM-X is a schema-validated envelope that ensures interoperability and safety across agents. Each message includes metadata (IDs, sender/recipient, timestamps) and a typed payload referencing a JSON schema. Below we show two illustrative sketches.

\paragraph{Envelope (generic).}
\begin{lstlisting}
{
  "envelope": {
    "msg_id": "uuid-1234",
    "ts": "2025-08-21T18:45Z",
    "from": "agent://alice",
    "to": ["topic://negotiation"],
    "capabilities": ["propose","negotiate"],
    "scope": {"consent": "opt-in", "ttl": 120}
  },
  "payload": { "$schema": ".../Negotiation.json", ... }
}
\end{lstlisting}

\paragraph{Example payloads.}
\begin{itemize}
    \item \textbf{CFP (Call-for-Proposal)}: includes \texttt{deadline}, \texttt{round}, and optional \texttt{constraints}.
    \item \textbf{Offer}: includes \texttt{value}, \texttt{timestamp}, and \texttt{conditions}.
    \item \textbf{Accept/Reject}: references the original \texttt{msg\_id} and contains the decision outcome.
    \item \textbf{Ack}: acknowledges delivery of a message, supporting at-least-once semantics.
\end{itemize}

While only sketches are provided here, full schemas for CFP, Offer, Accept/Reject, and Ack are made available in the repository to support replication and extension.

\section{Transport and Implementation Notes}
\label{appendix:transport}

The transport layer of LLM-X is deliberately lightweight, built on \textbf{NATS/HTTP} subjects such as \texttt{topic.*}, \texttt{agent.*.inbox}, and \texttt{ack.<id>}. Gateways enforce authentication (JWT), schema validation, and rate limiting before admitting messages. Once validated, messages are delivered to the negotiation substrate and regulated by the Policy Engine (Low, Medium, High).

Reliability is ensured by an \textbf{Ack+Retry} mechanism: if acknowledgments are not received within the configured TTL, retries are triggered with exponential backoff. This ensures at-least-once delivery while avoiding retry storms under load. Observability hooks record per-minute metrics, latency traces, and audit logs, which are exported to Prometheus-compatible collectors.

This design couples minimal substrate overhead with reproducibility: developers can swap brokers (e.g., NATS, Kafka) or observability backends without changing the negotiation logic.

\end{document}